\documentclass{article}

\usepackage[preprint,nonatbib]{neurips_2020}

\usepackage[utf8]{inputenc} 
\usepackage[T1]{fontenc}    
\usepackage{hyperref}       
\usepackage{url}            
\usepackage{booktabs}       
\usepackage{amsfonts}       
\usepackage{nicefrac}       
\usepackage{microtype}      
\usepackage{graphicx}
\usepackage{amsmath}
\usepackage{amssymb}
\usepackage{xcolor}
\usepackage{float}
\usepackage{multirow}
\usepackage{array}
\usepackage{booktabs}
\usepackage{marvosym}
\usepackage{fontawesome}
\usepackage{lipsum}

\newcommand{\filluptopage}[1]{%
  \clearpage
  \loop\ifnum\value{page}<#1\relax
    \null\clearpage
  \repeat
  \loop\ifnum\value{page}=#1\relax
    \null\clearpage
  \repeat
}

\newcommand{\eg}{\textit{e.g. }}
\newcommand{\ie}{\textit{i.e. }}

\title{Rethinking Sampling in 3D Point Cloud\\ Generative Adversarial Networks}

\vspace{-10mm}
\author{%
  He Wang$^{1}$\thanks{~denotes equal contributions.}\quad
  Zetian Jiang$^{2\ast}$\quad
  Li Yi$^{3}$\quad
  Kaichun Mo$^{1}$\quad
  Hao Su$^{4}$\quad
  Leonidas J. Guibas$^{1}$\\
  $^{1}$Stanford University, 
  $^{2}$Shanghai Jiao Tong Univeristy, \\
  $^{3}$Google Research, 
  $^{4}$University of California, San Diego\\
\vspace{-5mm}
}

\begin{document}
\maketitle

\begin{abstract}
\vspace{-2mm}

    In this paper, we examine the long-neglected yet important effects of point sampling patterns in point cloud GANs.
    Through extensive experiments, we show that sampling-insensitive discriminators (\eg PointNet-Max) produce shape point clouds with point clustering artifacts while sampling-oversensitive discriminators (\eg PointNet++, DGCNN) fail to guide valid shape generation.
    We propose the concept of sampling spectrum to depict the different sampling sensitivities of discriminators.
    We further study how different evaluation metrics weigh the sampling pattern against the geometry and propose several perceptual metrics forming a sampling spectrum of metrics.
    Guided by the proposed sampling spectrum, we discover a middle-point sampling-aware baseline discriminator, PointNet-Mix, which improves all existing point cloud generators by a large margin on sampling-related metrics.
    We point out that, though recent research has been focused on the generator design, the main bottleneck of point cloud GAN actually lies in the discriminator design. Our work provides both suggestions and tools for building future discriminators. We will release the code to facilitate future research.
\end{abstract}

\section{Introduction}
\vspace{-2mm}
Point cloud, as the most common form of 3D sensor data, has been widely used in a variety of 3D vision applications due to its compact yet expressive nature and its amenability to geometric manipulations. It is natural to consider how to generate point cloud through deep learning approaches, which has been a popular research topic recently. The previous research efforts in the community have been mainly devoted to conditional generation of point clouds with 3D supervision. The condition could either be images~\cite{fan2017point, groueix2018atlasnet, park2019deepsdf} or partial point clouds~\cite{li2019pu, yang2018foldingnet}. 
\begin{figure*}[th!]
\centering
  \includegraphics[width=\linewidth]{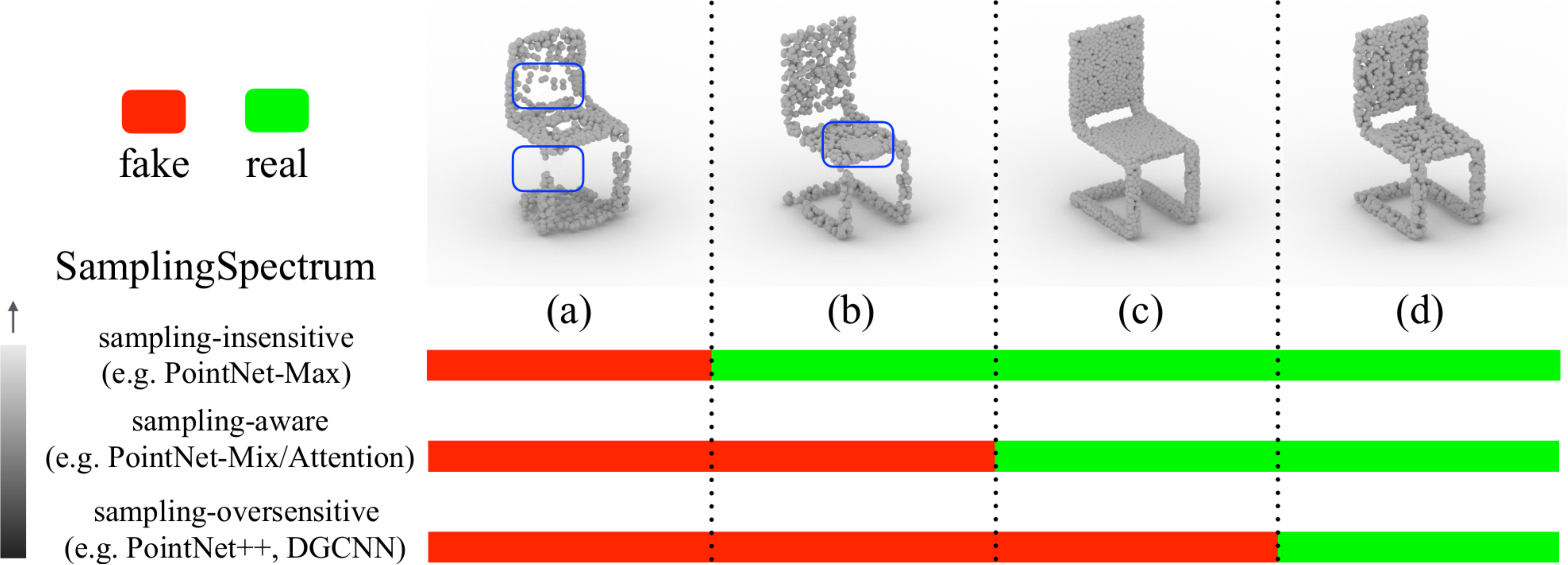}
  \vspace{-2mm}
  \caption{We visualize the behavior of different discriminators when judging four different chair point clouds. (a) and (b) are generated results, (c) and (d) are point clouds sampled using FPS and uniform sampling from a real chair surface. When training the discriminators on data like (d), different discriminators make distinct decisions on the point cloud realness forming a \textbf{sampling spectrum} ranging from sampling-insensitive, sampling-aware to sampling-oversensitive. 
  We advocate sampling-aware discriminators in the middle of the spectrum, which provide good guidance for fixing geometric flaws (a) and big sample artifacts (b) and tolerate subtle sampling differences between (c) and (d).
  }
  \vspace{-5mm}
  \label{fig:teaser}
\end{figure*}

Generating 3D point clouds with GANs in an unsupervised manner is an important but less explored problem. 3D point cloud GAN learns to transform a random latent code into a 3D surface point cloud by playing an adversarial game. Its development is still in an early stage compared with 2D image GANs.
While existing works such as \cite{achlioptas2017learning, valsesia2018learning, shu20193d} have developed a variety of generators, they all use PointNet~\cite{qi2017pointnet} with max pooling (PointNet-Max) as their discriminator. PointNet, which is essentially a pointwise MLP followed by a global pooling operation, is too limited in capturing shape details for a successful GAN. However, advanced networks, e.g. PointNet++\cite{qi2017pointnet++}, DGCNN\cite{wang2019dynamic}, which leverage relative positions between points and hierarchical feature extraction, may not help. From our empirical study, we find they both fail to be a functioning discriminator. Understanding their failure mechanism and further improving discriminator design are hence important and urgent.
 
To design a better discriminator, we first need to answer the following question: \textit{what should the discriminator examine for improving the generation quality?} Or, even more fundamentally, \textit{what does it mean by the quality of generated point clouds?}

Since a shape point cloud are the points sampled from an object surface, its quality should be evaluated from two perspectives: \textit{the depicted surface geometry} and \textit{the point sampling}. Arguably, geometry plays a decisive role and should be the main focus of a discriminator. However, 
when the generated point clouds have a good shape, there is still a full spectrum on how much a discriminator cares about the 
sampling patterns. 
We introduce the concept of \textbf{Sampling Spectrum} to depict the \textit{sampling sensitivity of discriminators}, as illustrated in Figure~\ref{fig:teaser}.
A sampling-insensitive discriminator (\eg PointNet-Max)
may ignore the point density variations as long as it perceives a good overall shape. Such a discriminator could identify big geometric flaws as shown in Figure~\ref{fig:teaser}~(a), but turns a blind eye to highly non-uniform density distribution, \eg point clusters in Figure~\ref{fig:teaser}~(b). 
On the other extreme, a sampling-oversensitive discriminator (\eg PointNet++, DGCNN) can even tell the subtle difference in sampling patterns, \eg between furthest point sampling (FPS) in Figure~\ref{fig:teaser}~(c) and uniform sampling in Figure~\ref{fig:teaser}~(d), 
and hence can be very narrow-minded about what a real point cloud should look like.
A sampling-aware discriminator (\eg PointNet-Mix/Attention), which lies in the middle of the spectrum, is able to identify density-related artifacts such as point clusters in Figure~\ref{fig:teaser}~(b),
while not being too sensitive to different sampling patterns, such as Figure~\ref{fig:teaser}~(c)~and~(d).

Resembling the sampling spectrum of discriminators, we also examine the existing point-cloud GAN evaluation metrics from the perspective of sampling sensitivity and propose several perceptual metrics forming a \textbf{Sampling Spectrum} of \textit{evaluation metrics}. Understanding how the metrics weigh between sampling and geometry is a prerequisite for evaluating point cloud GANs.
Many of the existing sampling-insensitive metrics only evaluate the geometry factor of the generated point clouds shapes, which are blind to the obvious point clustering artifacts and uneven point density.
We propose novel sampling-sensitive metrics to further complete the spectrum of point-cloud GAN evaluation metrics.

Guided by the proposed sampling spectrum of discriminators and evaluation metrics, experiments show that different discriminators in the spectrum could provide very different suggestions to improve a generator according to its sampling sensitivity.
%
Sampling-insensitive discriminators (\eg PointNet-Max) is unaware of point density variations and hence its generated point cloud inevitably suffer from clustering artifacts, 
while sampling-oversensitive discriminators (\eg PointNet++~\cite{qi2017pointnet++} and DGCNN~\cite{wang2019dynamic}) simply fail to function as the discriminators and can generate much degraded point cloud shapes.
We design a diagnostic ``no generator'' experiment to factor out the impact from generators and reveal that the gradients of sampling-oversensitive discriminators prioritize adjusting sampling patterns over producing better shape geometry.
Picking a middle-point on the sampling spectrum, we discover a simple yet effective sampling-aware discriminator, PointNet-Mix, and find that it can supervise both shape generation and point density uniformity.
It improves all existing generators by a large margin on sampling-related metrics. 
Surprisingly, we find that even the most naive fully-connected generator, coupled with PointNet-Mix,
simply beats all the start-of-the-art point cloud GANs.
This discovery conveys an important message to the community: instead of focusing on the generator design, people should invest more time into discriminator and seek for more powerful sampling-aware discriminators.

\vspace{-2mm}
\section{Point Cloud GAN Landscape}
\label{sec:background}
\vspace{-2mm}
In this section, we review the current state of point cloud GAN covering the generators, the discriminators, and the evaluation metrics we are examining in this work. 

\vspace{-2mm}
\subsection{Point Cloud GAN Generators}
\vspace{-2mm}
Recent point cloud GAN works primarily focus on the generator design. The existing generators can be categorized into two classes: fully-connected (FC) generators and graph convolutional generators. The first point cloud GAN, r-GAN~\cite{achlioptas2017learning}, simply uses an FC network as its generator. GraphCNN-GAN~\cite{valsesia2018learning} and TreeGAN~\cite{shu20193d} are the rest two published works in this field that use graph convolutional generators. The two methods are very similar in principle. The main difference lies in how they build graphs. GraphCNN-GAN builds a dynamic $k$-nn graph based upon feature space distance while TreeGAN enforces a tree structure throughout its sequential graph expansion and the messages can only be passed from ancestors vertices to descendants vertices. 

Deformation-based decoders~\cite{groueix2018atlasnet,yang2018foldingnet} are widely used in the point cloud auto-encoder networks for 3D shape reconstruction. The decoders leverage Multiple-layer Perceptrons (MLP) to deform template surfaces into shape surfaces taking as inputs the concatenation of template point coordinates and the latent shape feature vectors. Though the decoders can truly act as generators for point cloud GANs, they have not yet been used in unconditioned point cloud GAN literature. Recently, Mo~et~al.~\cite{mo2020pt2pc} use deformation-based decoder as part generators for structure-conditioned point cloud GAN.
\vspace{-2mm}
\subsection{Point Cloud GAN Discriminators}
\vspace{-2mm}
\label{sec:discriminator}
All existing works on unconditioned point cloud GANs use PointNet~\cite{qi2017pointnet} with max-pooling (PointNet-Max) as their discriminators. 
PointNet learns a function $h$ that maps each point $p_i$ in the point cloud to a per-point feature $h(p_i)\in \mathbb{R}^d$ and then extracts a permutation-invariant global feature $F \in \mathbb{R}^d$ by pooling the per-point features across all points using a symmetric function $g$, which can be max pooling, average pooling, etc. Namely, we have $F=g(\{h(p_1), h(p_2), \cdots, h(p_N)\})$.

\vspace{-3mm}
\paragraph{PointNet-Max/Avg.}
Though $g$ can be any symmetric function, most existing works use max-pooling.
In PointNet~\cite{qi2017pointnet}, the authors show that max-pooling outperforms average-pooling on 3D shape classification tasks.
They further show that the global feature $F$ obtained from max-pooling is determined by only a sparse subset of the points, namely critical points $\mathcal{C}_S$, 
bringing PointNet-Max with robustness against small data perturbation and corruption. 
However, this property may limit its discriminative power on telling the density variations and classifying different sampling patterns. 
To investigate how different aggregation operations affect the sampling sensitivity and generation quality, we study the two common choices of the symmetric function $g$, max-pooling, and average-pooling, in this paper.
We will show in Sec.~\ref{sec:sampling_sensitivity} that using different aggregation operations makes huge differences when adapting PointNet-based networks as point cloud GAN discriminators.

\vspace{-3mm}
\paragraph{PointNet-Mix.}
By simply concatenating the max-pooling feature and the average pool feature, we obtain another permutation-invariant feature. We name this PointNet-Mix. 
Formally, $F_\text{mix} = [\text{max}\{h(p_1), ..., h(p_N)\}; \text{avg}\{h(p_1), ..., h(p_N)\}] \in \mathbb{R}^{2d}.$
The mix-pooling operation is a special choice of the symmetric function $g$.
We will show in our experiments that PointNet-Mix, though simple, surprisingly improves the performance for most point cloud GANs by a large margin on sampling-aware metrics.


\vspace{-3mm}
\paragraph{PointNet-Attention.}
A recent point cloud upsampling work~\cite{li2019pu} incorporates a self-attention module into PointNet for its discriminator and shows improved point density in the upsampled point cloud. We denote this discriminator as PointNet-Attention. 
The self-attention module learns three separate MLPs to transform each $h(p_i)$ into $f(p_i), l(p_i), k(p_i) \in \mathbb{R}^{d}$ correspondingly. 
Then an attention weight matrix is formed by $W = \text{SoftMax}\left([f(p_1), ..., f(p_N)]^{T}[(l(p_1), ..., l(p_N))]\right) \in \mathbb{R}^{N\times N}$. 
A weighted features is then obtained through $w(p_i) = h(p_i) + W^{T}k(p_i)$. 
The final aggregated features of PointNet-Attention is a max-pooling of the weighted feature, namely $F_\text{attention} = \text{max}\{w(p_1), ..., w(p_N)\}$.
Note that PointNet-Attention allows the per-point features to communicate with each other according to their similarity, which is more sensitive to sampling patterns than PointNet-Max.
We investigate this strategy in our paper as well.

\vspace{-3mm}
\paragraph{Discriminators beyond PointNet.}
Recently, there have been many works~\cite{qi2017pointnet, li2018pointcnn, wang2019dynamic, hermosilla2018monte, thomas2019kpconv} extending PointNet to more advanced 3D deep learning architectures on point clouds. 
They improve PointNet by extracting more local or hierarchical geometric features via point cloud convolutions or point cloud graph learning.
Though proven to be effective on shape classification and segmentation tasks, no published work examines adapting them as point cloud GAN discriminators.
In this paper, we investigate two exemplar beyond-PointNet discriminators: PointNet++~\cite{qi2017pointnet++} and DGCNN~\cite{wang2019dynamic}.

\vspace{-2mm}
\subsection{Point Cloud GAN Evaluation Metrics}
\vspace{-2mm}
Achlioptas~et~al.~\cite{achlioptas2017learning} introduces two distance metrics in the Euclidean space for evaluating point cloud GAN: the coverage scores (COV) computing the fraction of the point clouds in $B$ that are closed to the point clouds in $A$ using either Earth Mover's distance~(EMD) or Chamfer distance~(CD), and the Minimum matching distance (MMD) scores measuring the fidelity of $A$ with respect to $B$ using either EMD or CD. 
In the field of 2D image GAN, it is common to use perceptual metrics, such as the Frech\'et distance~\cite{heusel2017gans} between real and fake Gaussian measures in the feature spaces, for evaluating the generated image results. Formally, $\text{Frech{e}t Distance} = ||\mu_r - \mu_g||^2 + \text{Tr}(\Sigma_r + \Sigma_g - 2\left(\Sigma_r\Sigma_g\right)^{1/2}),$
where $\mu$ and $\Sigma$ are the mean vector and the covariance matrix of the features calculated from either real or fake data distribution, and $\text{Tr}$ is the matrix trace. 
For point clouds, Shu~et~al.~\cite{shu20193d} proposes Frech\'et Point Cloud Distance (FPD), which uses the features extracted from a pre-trained PointNet-Max model. 
In this paper, we position the existing metrics on a sampling spectrum according to their sampling sensitivity and propose novel sampling-aware metrics to augment the spectrum.

\vspace{-2mm}
\section{Sampling Spectrum of Discriminators}
\label{sec:sampling_sensitivity}
\vspace{-2mm}
While recent works propose many advanced generator improvements for point cloud GANs, 
we find that designing a good discriminator is of equal importance, if not more.
In this section, we introduce \textit{the sampling spectrum of discriminators}, on which we thoroughly examine the positions of different discriminators that explain their behaviors when training point cloud GANs.

\vspace{-2mm}
\subsection{Sampling Sensitivity of Discriminators}
\vspace{-2mm}
The sampling sensitivity of a discriminator depicts how much it responds to a change in the point density or sampling pattern of an input point cloud. We find it extremely hard to quantitatively measure this sensitivity given the difficulty of measuring changes in the sampling patterns. Naively using Euclidean metrics (\eg CD or EMD) to measure the distance between two sampled point sets is not a solution, since given the same distance budget, the discriminator's responses can be dramatically different depending on how the sampled points move. Instead, we can set landmarks in the continuous spectrum of sampling sensitivity by examining the discriminative power of the discriminators against different sampling patterns under a series of experiments from easy to hard. Specifically, we design two experiments to test whether a discriminator could tell clustering artifacts in point clouds and whether it could distinguish between FPS and uniform sampling patterns. Accordingly, we divide the spectrum into three regimes: sampling-insensitive, sampling-aware, and sampling-oversensitive, as shown in Fig.\ref{fig:teaser}.

\textbf{A sampling-insensitive discriminator} does not respond to local point density changes if the overall shape remains roughly the same. This kind of discriminators can't tell clustering artifacts in the generated point clouds, i.e. Fig.\ref{fig:teaser} (b), and thus may cause the non-uniform density in the generated point clouds. 

\textbf{A sampling-aware discriminator} can notice the significant point non-uniformity and is hence capable of supervising the generator to enforce a similar sampling distribution to the training data, while being ignorant to subtle changes when the sampling is already uniform in an intermediate scale. Such discriminator will judge Fig.\ref{fig:teaser} (b) as fake but can't tell the difference between Fig.\ref{fig:teaser} (c) and (d).

\textbf{A sampling-oversensitive discriminator} can tell very subtle changes in the sampling patterns, \eg the difference between FPS and uniform sampling even if the shape of two point clouds are the same (Fig.\ref{fig:teaser} (c) and (d)).


\vspace{-2mm}
\subsection{Sampling Sensitivity Examination Results}\label{sec:disc_sampling_sensitivity}
\vspace{-2mm}
We design diagnostic experiments to categorize point cloud GAN discriminators on the sampling spectrum: sampling-insensitive (PointNet-Max~\cite{qi2017pointnet}), sampling-aware (PointNet-Avg~\cite{qi2017pointnet}, PointNet-Mix,  PointNet-Attention~\cite{li2019pu}), and sampling-oversensitive (PointNet++~\cite{qi2017pointnet++} and DGCNN~\cite{wang2019dynamic}).


\begin{table}[t!]
  \centering
  \scriptsize
  \begin{tabular}{c|c|c|c|c|c|c}
    \toprule
    {Experiment} & {PointNet-Max} & {PointNet-Avg} & {PointNet-Mix} & {PointNet-Attention} & {PointNet++} & {DGCNN} \\ \hline
    {Clustering artifacts}  & 56\%/53\% & 98\%/99\%  & 97\%/96\% & 94\%/93\% & 100\%/98\% & 100\%/99\% \\
    {FPS vs. uniform}  & 50\%/50\% & 50.3\%/50\%  & 51.5\%/50\% & 50.1\%/50\% & 100\%/97\% & 100\%/96\%\\ 
    \bottomrule
  \end{tabular}
  \caption{Evaluating the discriminating power of the discriminators against clustering artifacts and sutble change in sampling patterns. In each cell, the number on the left show the training accuracy while the number of the right shows the test accuracy.}
  \label{table:density_classification}
  \vspace{-7mm}
\end{table}

\vspace{-2mm}
\paragraph{The Discriminating Power against Clustering Artifacts.}
\label{sec:density_classification}
\vspace{-2mm}
To examine whether the discriminators can tell clustering artifacts or not, we create a diagnostic classification dataset from ShapeNet~\cite{chang2015shapenet}. Taking 100 shapes from the chair class, we uniformly sample 2048 points from each shape, forming a set of real point clouds. To form a fake set of point clouds, for each chair, we first uniformly sample 1024 points, and then densely sample another 1024 points around a random position on the chair within a 0.1 radius.
The real/fake point clouds are used as the labeled training dataset. We repeat the same process to generate a test dataset using a different set of 100 chairs. 

We supervisely train each discriminator to classify the real/fake point clouds. We train them 200 epochs until convergence. The training and test accuracies are shown in the first row of Table~\ref{table:density_classification}. Despite the huge density variation and the remarkable clustering artifacts, PointNet-Max is just barely better than a random guess while the rest of the discriminators are very successful in telling the fake from the real. This indicates that \textit{PointNet-Max is sampling-insensitive}. Note that the discriminating power of a network towards certain artifacts is maximized under such supervised training scheme. A network will fail to identify such artifacts in an adversarial training scheme if it fails in a supervised training scheme. We will see in Sec.~\ref{sec:exp}, point clouds generated by GAN using PointNet-Max as the discriminator indeed suffer from non-uniform density artifacts.

Our insight why PointNet-Avg/Mix can tell the artifacts but PointNet-Max fails is that the average pooling feature computes the center of the mass of the points in feature space and is hence aware to certain global non-uniform density distributions. PointNet-Attention leverages a learnable weighted averaging and is hence capable to identify the difference.

\vspace{-2mm}
\paragraph{Distinguishing between FPS and Uniform Sampling.}
We construct another diagnostic dataset with real and fake data which are the same in their shapes but only differ in their sampling patterns. Specifically, we perform uniform sampling to generate real data and use FPS to generate fake data from 100 chairs. Figure~\ref{fig:teaser} (c) and (d) illustrate the different sampling pattern outcomes.

We present the training and test accuracies in the second row of Table~\ref{table:density_classification} and show that both PointNet++ and DGCNN can perfectly distinguish the subtle difference in sampling patterns while the rest discriminators make no progress even on the training set. The experiment indicates that \textit{PointNet-Avg/Mix/Attention are sampling-aware} while \textit{PointNet++ and DGCNN are sampling-oversensitive}. 

We believe, for PointNet++ and DGCNN, their capability of distinguishing FPS from uniform sampling owes to their usage of relative point positions or edge information, which are highly sensitive to any change in sampling. We will show in Sec.~\ref{sec:eval_pointnet2_dgcnn} that this remarkable discriminating power of PointNet++ and DGCNN on sampling patterns actually leads to their failures as a functioning discriminator to train point cloud GANs.

\vspace{-2mm}
\section{Sampling Spectrum of Evaluation Metrics}
\label{sec:metric}
\vspace{-2mm}
Similar to the discriminator design, it is very important to understand how different evaluation metrics weigh the differences in sampling patterns against geometry quality.
Thus, we introduce \textit{the sampling spectrum of evaluation metrics}, which exactly resembles the sampling spectrum of discriminators introduced in Sec.~\ref{sec:sampling_sensitivity}. 
On the spectrum, we have sampling-insensitive metrics that measure only the shape geometry and are ignorant of the sampling patterns, and sampling-sensitive metrics that measure both at the same time.

For perceptual metrics, \ie Frech\'et distances in feature spaces, the sampling sensitivity of the metric purely depends on the sampling sensitivity of its feature extractor. Frech\'et distance measured in different feature spaces may respond very differently to changes in point density and sampling patterns. 
In this work, we examine three Frech\'et distance metrics that extract features using PointNet-Max, PointNet-Mix, and DGCNN, respectively. We denote them as  Frech\'et PointNet-Max Distance (FPD-Max), Frech\'et PointNet-Mix Distance (FPD-Mix), and Frech\'et DGCNN Distance (FGD). We pretrain all the three feature extraction networks on ModelNet40~\cite{wu20153d} shape classification. 

To examine the sampling sensitivity of FPD-Mix/Max and FGD, we create several copies of the training split of our ShapeNet chair dataset (see Sec.\ref{sec:dataset}), each of which uses a different sampling strategy to obtain the shape point clouds. The reference one used as the ground truth is using uniform sampling. Then we consider 1) uniform sampling with a different random seed; 2) FPS; and 3) biased sampling with clustered artifacts (as described in Sec.~\ref{sec:density_classification}). We use all the available metrics to evaluate their distances to the ground truth data. The results are shown in Table \ref{table:metrics_results}.

We observe that \textit{the Frech\'et distance metrics share the same sampling sensitivity of their corresponding discriminators}. 
For example, since PointNet-Max is sampling-insensitive, FPD-Max remains very low even on biased sampling data, hence FPD-Max serves as a perceptual geometry metric, which is ignorant to sampling patterns. 
Similarly, we find that FPD-Mix is sampling-aware since it clearly detects the biased sampling patterns while not being able to distinguish the uniform sampling and FPS,
while FGD is sampling-oversensitive in that it can tell apart FPS and uniform sampling.

For Euclidean distance metrics, results in Table \ref{table:metrics_results} show that COV-EMD and MMD-EMD are sampling-aware, which is intuitively reasonable since EMD enforces a one-to-one matching and is aware of the point density, while COV-CD and MMD-CD are sampling-insensitive.

\begin{table*}[t!]
  \centering
  \scriptsize
  \begin{tabular}{c|c|c|c|c|c|c|c}
    \toprule
    {Data} & {FPD-Mix $\downarrow$} & { FPD-Max$\downarrow$} & {FGD$\downarrow$}   & { MMD-E$\downarrow$} &  { MMD-C$\downarrow$} & {COV-E$\uparrow$} & {COV-C$\uparrow$}\\\hline
    Uniform re-sampling & 0.1153 & 0.0926 & 0.8141 & 0.1104 & 0.00145 & 70.69 & 72.16\\
    Farthest point sampling & 0.1700 & 0.1558 & 1.8833 & 0.1064 & 0.00137 & 67.74 & 69.36 \\
    Biased sampling & 2.8631 & 0.3524 & 9.6719 & 0.2469 & 0.00145 & 23.12 & 71.28\\
    \bottomrule
  \end{tabular}
  \caption{Examining sampling sensitivity of evaluation metrics.}
  \label{table:metrics_results}
  \vspace{-3mm}
\end{table*}


\vspace{-2mm}
\section{Experiments}
\label{sec:result}
\vspace{-2mm}
Aware of both the sampling spectrums, we conduct experiments to further evaluate the performance of point cloud GANs under various evaluation metrics.
We show that the point cloud GANs using sampling-insensitive discriminators may produce point clustering artifacts, while sampling-oversensitive discriminators fail to supervise point cloud GAN training at all.
We further devise a diagnostic "no-generator" experiment that factors out the generators to better illustrate our discoveries on discriminators.
More interesting, we find that the simple PointNet-Mix paired of any generator, even with the most naive fully-connected one, achieves the state-of-the-art performance.

\vspace{-2mm}
\subsection{Setting and Datasets}\label{sec:dataset}
\vspace{-2mm}
We provide a thorough comparison of all the discriminators investigated in Sec.\ref{sec:sampling_sensitivity} combining with all the available generators in the published literature, including the FC generator proposed in r-GAN~\cite{achlioptas2017learning}, and graph convolutional generators used in TreeGAN~\cite{shu20193d}. We also add a deformation-based generator into the comparison given its popularity for supervised point cloud reconstruction~\cite{groueix2018atlasnet, yang2018foldingnet}. 

We use two datasets to evaluate the GANs. One is a single-category dataset containing point clouds sampled from all 6,778 chair meshes in ShapeNet~\cite{chang2015shapenet}. The other is a multi-category dataset combining shapes from airplane, car, chair, rifle, sofa, table, vessel categories in ShapeNet. The multi-category dataset contains 34,313 shapes in total. We uniformly sample 2048 points from each shape to form the two datasets.
We follow the 85\%/5\%/10\% train/validation/test split in~\cite{achlioptas2017learning} and use WGAN-gp~\cite{arjovsky2017wasserstein,gulrajani2017improved} for the GAN training, similar to previous works~\cite{achlioptas2017learning,valsesia2018learning,shu20193d}.

\vspace{-2mm}
\subsection{Evaluating PointNet-based Discriminators with Various Generators}
\vspace{-2mm}
We report the performance for point cloud GANs that combine PointNet-Max/Min/Attention discriminators and FC/Deform/TreeGAN/Graph-CNN generators in Table~\ref{table:results}.
We observe that GANs using PointNet-Mix as the discriminator outperform the ones using PointNet-Max/Attention across all different generators. 
On sampling-aware/sensitive metrics (\ie FPD-Mix, FGD, MMD-EMD, COV-EMD), PointNet-Mix is always significantly better than PointNet-Max and is better than PointNet-Attention mainly on FPD-Mix and FGD. Regarding geometry quality evaluated using sampling-insensitive metrics (\ie FPD-Max, MMD-CD, COV-CD), PointNet-Mix is always significantly better than PointNet-Attention on FPD-Max and COV-CD while remaining on a par with PointNet-Max. 

\begin{figure}[t!]
\centering
  \centering
  \includegraphics[width=\linewidth]{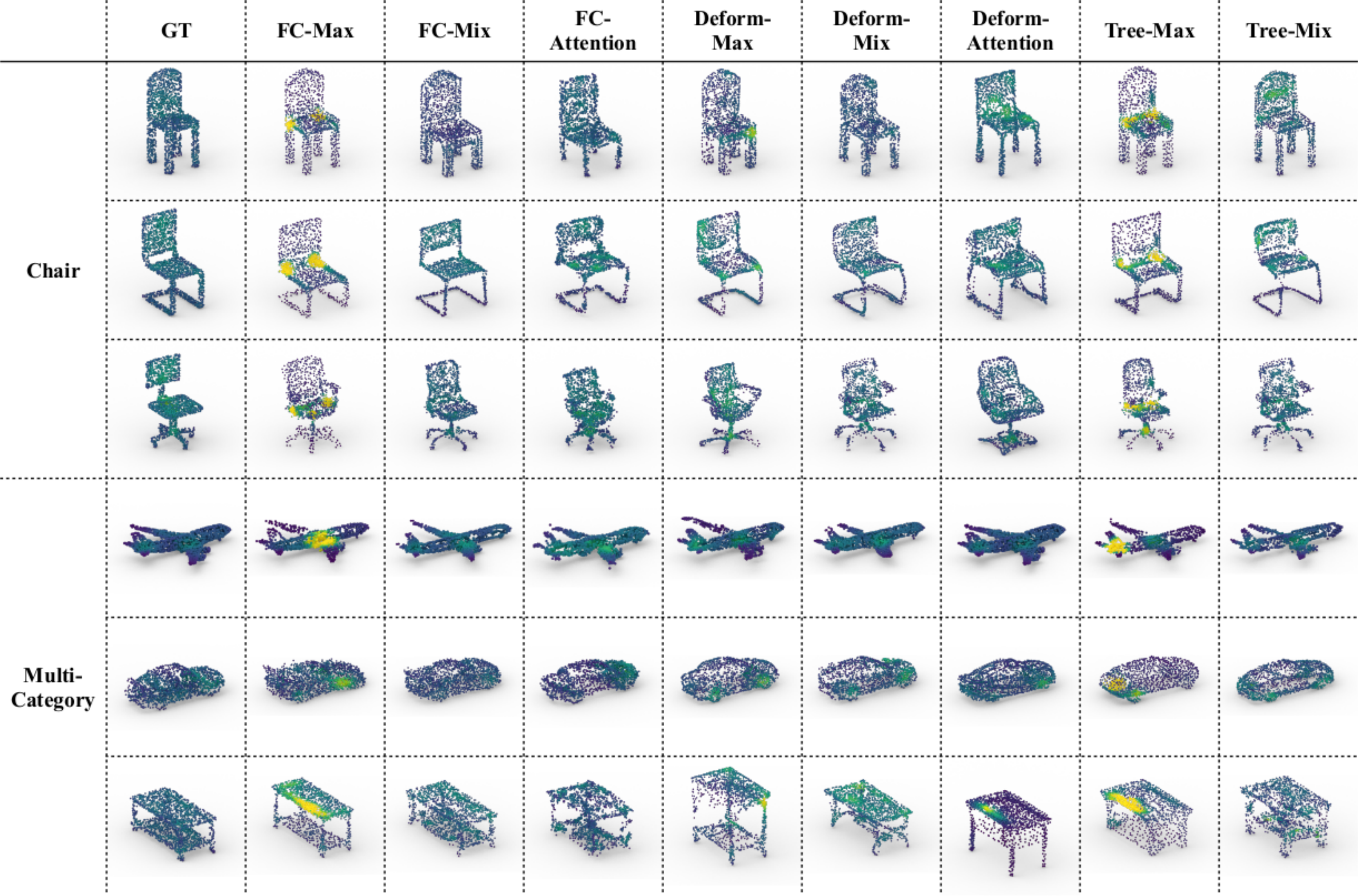}
     \vspace{-5mm}
  \caption{\textbf{Visualization of point clouds generated by different methods.} 
  We show generated point clouds in few exemplar shapes for a fair and easier comparison regarding their sampling quality and geometry quality (the point clouds are still random selected without cherry-picking).
  We color-code the local point density that ranges from sparse (dark blue) to dense (light yellow).}
  \label{fig:generation}
   \vspace{-5mm}
\end{figure}

In Figure~\ref{fig:generation}, we present the generated point clouds for all the experiments with color-coding for the local point density. 
We see that the generators trained using PointNet-Max usually suffer from non-uniform density, except for the deformation generator. On the chair class, points are usually clustering around the seat area while being sparse at the back. On the contrary, PointNet-Mix enforces a globally uniform point density and hence greatly improves the visual quality of the generated point clouds. PointNet-Attention is in the between. 

With PointNet-Mix, we observe that the most naive FC generator works the best outperforming the previously state-of-the-state generators on almost of metrics. This showcases the importance of being sampling-aware but not sampling-insensitive/oversensitive as a discriminator. It also suggests that future works may focus on designing more powerful sampling-aware discriminators.
\label{sec:exp}
\begin{table*}[t!]
  \centering
  \scriptsize
  \begin{tabular}{c|c|c|c|c|c|c|c|c|c}
    \toprule
    {Dataset} & {Generator} & {Pooling} & {FPD-Mix $\downarrow$} & { FPD-Max$\downarrow$} & {FGD$\downarrow$}   & { MMD-E$\downarrow$} &  { MMD-C$\downarrow$} & {COV-E$\uparrow$} & {COV-C$\uparrow$}\\\hline
    \multirow{10}{*}{{Chair}}
    & {FC}             & Max       & 1.571 & 0.211  & 7.030 & 0.1017 & 0.00164 & 23.56 & 72.75\\
    & {FC}             & Mix       & \textbf{0.184} & \textbf{0.209}  & \textbf{2.124} & \textbf{0.0674} & 0.00196 & 73.64 & 74.96\\
    & {FC}             & Attention & 0.635 & 0.672  & 4.971 & 0.1156 & \textbf{0.00160} & 68.92 & 70.54 \\
    & {Deform}    & Max       & 0.913 & 0.268  & 5.602 & 0.0908 & 0.00201 & 68.5 & 72.61 \\ 
    & {Deform}    & Mix       & 0.534 & 0.373  & 2.836 & 0.0695 & 0.00200 & \textbf{76.29} & 75.11\\ 
    & {Deform}    & Attention & 0.696 & 0.755  & 2.987 & 0.1141 & \textbf{0.00160} & 68.77 & 69.36 \\
    & {TreeGAN}        & Max       & 1.442 & 0.654  & 7.808 & 0.0962 & 0.00191 & 24,74 & 73.49\\
    & {TreeGAN}        & Mix       & 0.293 & 0.334  & 4.032 & 0.0704 & 0.00211 & 74.82 & \textbf{78.79}\\
    & {Graph-CNN}      & Max       & 1.034 & 0.981 & 7.494 & 0.0812 & 0.00191 & 48.90 & 63.18\\
    \midrule
    \multirow{9}{*}{{Multi}}
    & {FC}          & Max       & 1.553 & 0.354 & 6.981 & 0.0842 & 0.00153 & 35.16 & 64.16\\ 
    & {FC}          & Mix       & \textbf{0.255} & \textbf{0.285} & 2.550 & \textbf{0.0656} & 0.00184 & \textbf{73.5} & \textbf{72.16} \\
    & {FC}          & Attention & 0.414 & 0.453 & 4.234 & 0.1188 & \textbf{0.00134} & 72.19 & 69.63 \\
    & {Deform} & Max       & 1.072 & 0.633 & 3.845 & 0.0799 & 0.00179 & 62.5 & 64.5\\ 
    & {Deform} & Mix       & 0.614 & 0.349 & \textbf{2.451} & 0.0670 & 0.00191 & 70.83 & 68.83\\ 
    -Cat& {Deform} & Attention & 0.616 & 0.720 & 2.531 & 0.1113 & 0.00141 & 72.04 & 69.60 \\
    & {TreeGAN}     & Max       & 1.714 & 0.437 & 6.342 & 0.1093 & 0.00158 & 25.33 & 67.0\\
    & {TreeGAN}     & Mix       & 0.388 & 0.420 & 4.300 & 0.0699 & 0.00184 & 72.66 & 71.0\\
    \hline
  \end{tabular}
  \vspace{-3mm}
  \caption{\textbf{Evaluating PointNet-based discriminators with different generators.} We observe that PointNet-Mix can significantly improve all sampling-aware/sensitive metric, including FPD-Mix, FGD, MMD-EMD, COV-EMD for all the generators. When paired with FC generator, PointNet-Mix achieves the best performance. }
  \label{table:results}
  \vspace{-3mm}
\end{table*}


\vspace{-2mm}
\subsection{Diagnosing Failures for PointNet-Avg, PointNet++ and DGCNN Discriminators}
\label{sec:eval_pointnet2_dgcnn}
\vspace{-2mm}
Observing that the discriminator forms the bottleneck of the point cloud GAN, it is important to examine whether advanced networks, \eg PointNet++~\cite{qi2017pointnet++}, DGCNN~\cite{wang2019dynamic}, can outperform PointNet-based discriminator. From our extensive experiments, we fail to use any of them as a discriminator to train a good GAN with any existing generators. 
During GAN training, we observe the two networks behave very similarly: they are discriminative in the sense that the gaps between the real and fake scores remain significantly large, but the quality of the generated point clouds stays very bad.
%
The contradictory behaviors seem unreasonable at the first glance: 
how can a discriminator be very discriminative but teach nothing to the generator? 
Thus, we design a diagnostic "no generator" experiment to examine the underlying reasons.

\begin{figure}[t!]
\centering
    \vspace{-4mm}
    \begin{minipage}[b]{0.4\columnwidth}
        \includegraphics[width=1.0\columnwidth]{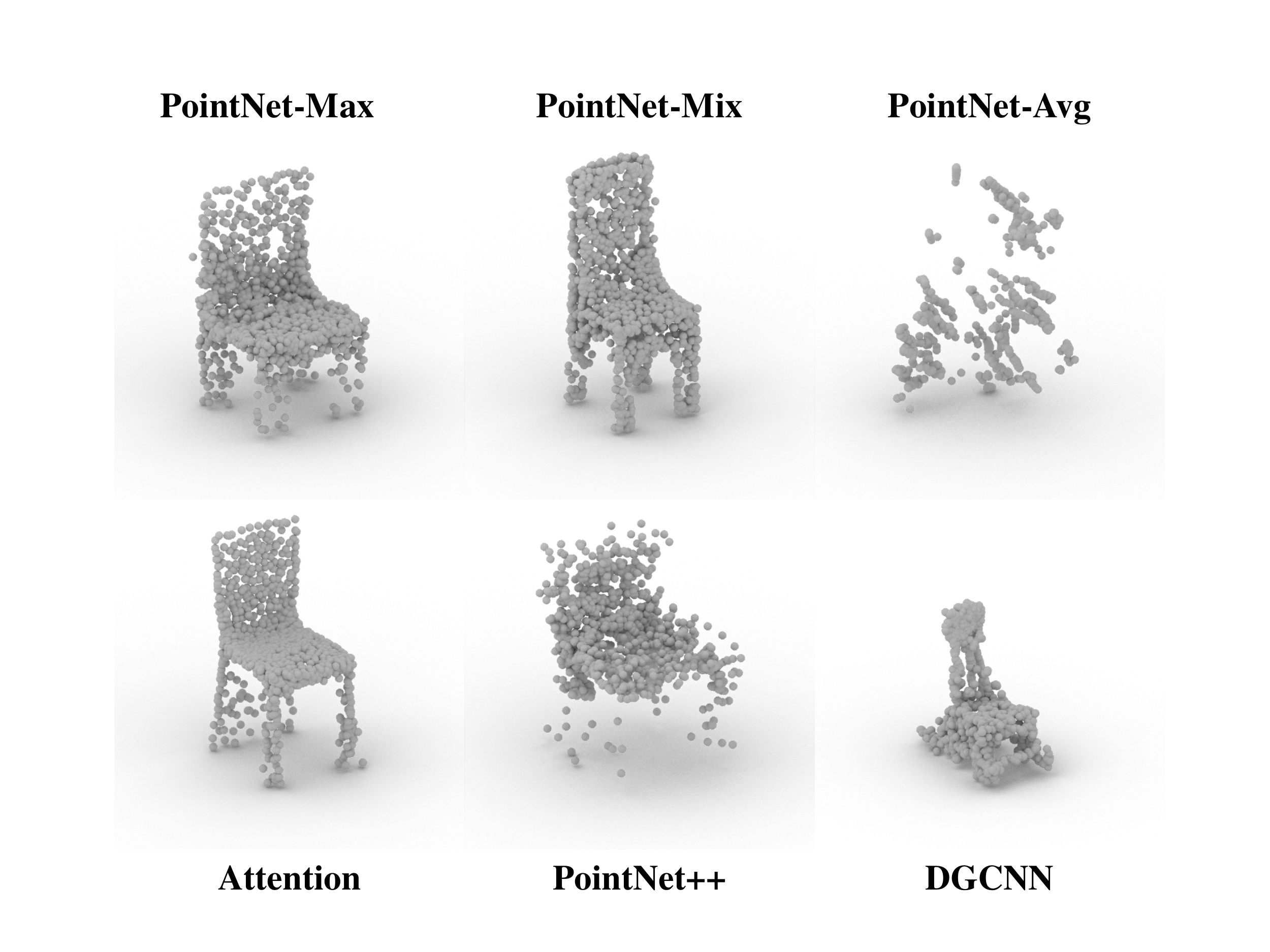}
        \label{fig:no_generator_result}
    \end{minipage}
    \begin{minipage}[b]{0.55\columnwidth}
        \includegraphics[width=\columnwidth]{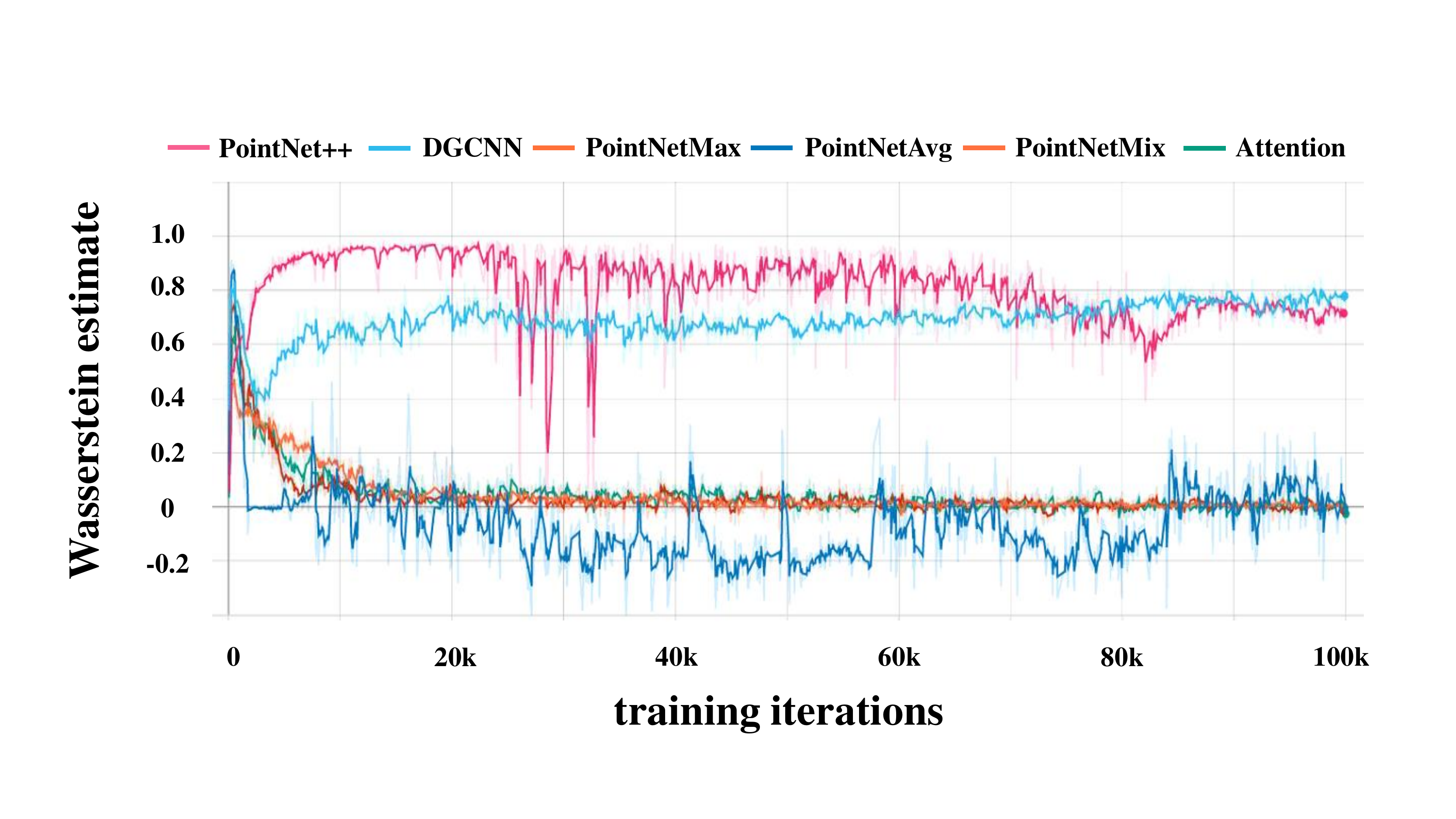}
        \label{fig:no_generator_train_curve}
    \end{minipage}
    \vspace{-5mm}
    \caption{\textbf{The diagnostic "no generator" experiment results}. On the left, we show exemplar point clouds generated by different discriminators. On the right, we plot the training curves for the experiments. The $x$- and $y$-axis respectively represent training iterations and Wasserstein estimates.}
    \vspace{-3mm}
    \label{fig:no_gen_figure}
\end{figure}

\begin{table}[t!]
  \centering
  \scriptsize
  \begin{tabular}{c|c|c|c|c|c|c|c}
    \toprule
     {Discriminator} & {FPD-Mix $\downarrow$} & {FPD-Max$\downarrow$} & {\hspace{6pt }FGD$\downarrow$ \hspace{5pt}}   & {MMD-E$\downarrow$} &  {MMD-C$\downarrow$} & {COV-E$\uparrow$} & {COV-C$\uparrow$}\\\hline
     {PointNet-Max} & 3.828 & 2.456 & 19.279  & 0.1039 & 0.00287 & 26 & \textbf{56} \\
     {PointNet-Avg} & 19.640 & 19.604 & 39.793 & 0.1154 & 0.00642 & 32 & 12  \\
     {PointNet-Mix} & 1.835 & 2.208 & 13.197 & \textbf{0.0898} & 0.00331 & 50 & 51  \\
     {PointNet-Attention} & \textbf{1.350} & \textbf{1.642} & \textbf{12.359} & 0.1622 & \textbf{0.00271} & \textbf{52} & 48  \\
     {PointNet++} & 15.081 & 14.187 & 33.958 & 0.1155 & 0.00503 & 19 & 37  \\
     {DGCNN} & 10.629 & 9.939 & 14.882 & 0.0932 & 0.00396 & 46 & 48\\ \bottomrule
    \end{tabular}
  \caption{\textbf{Quantitative evaluation for the diagnostic "no generator" experiments}. We see that PointNet-Max/Min/Attention are successful while PointNet-Avg/PointNet++/DGCNN fail.}
  \label{table:nogen_results}
  \vspace{-5mm}
\end{table}

\paragraph{No Generator Experiments.}
\vspace{-2mm}
During GAN training, the gradients from discriminator output scores back-propagate to the generated point clouds and then further back-propagate to the weights of the generator to supervised the generator training.
We design a diagnostic experiment that removes the generator and examines whether the discriminator gradients are informative enough for supervising point cloud GANs.
Concretely, we randomly initialize a set of learnable point clouds with i.i.d Gaussian noises $\mathcal{N}(0, 0.1)$.
Then, we conduct adversarial training where the discriminator gradients directly update the learnable point clouds.
%
The real data contains point clouds with 2,048 points uniformly sampled from 100 chairs. We train each experiment for 10,000 epoches until convergence. 

\paragraph{Results and Analysis.}
\vspace{-2mm}
Table~\ref{table:nogen_results} summarizes the results and Figure~\ref{fig:no_gen_figure} (left) shows exemplar generated point clouds. 
We clearly see that
only PointNet-Max/Mix/Attention can successfully modify the learnable point clouds to get high quality shapes, 
while PointNet++/DGCNN produce much worse results and PointNet-Avg are the worst from which one can only barely identify any object.
%
%
Figure~\ref{fig:no_gen_figure} (right) presents the training curves of the Wasserstein estimates, which essentially describe the gap between the real and fake scores. 
The large score gaps for PointNet++ and DGCNN throughout the training indicate the two models are very discriminative in telling apart the fake samples from the real data. However, their gradients simply don't help improve the learned point clouds. Note that both of them leverage relative point positions/edge information during their feature extraction, which leads to a huge amount of gradients flowing along the surface direction focusing on changing sampling patterns instead of supervising shape geometry. 
For PointNet-Avg, the failure is simply because the discriminating power is not sufficient, evident from the low Wasserstein estimates.
Note that training such "no generator" experiment shares a similar flavor to the SGD sampling in introspective CNN\cite{jin2017introspective}.

\vspace{-2mm}
\section{Conclusion and Suggestions for Future Discriminator Design}
\vspace{-2mm}

In this work, we study the importance of sampling for 3D point cloud GAN design.
We propose the sampling spectrum of discriminators and evaluation metrics that provide insights on the behaviors of point cloud GANs.
We propose several empirical experiments for identifying the sampling sensitivity of a discriminator or an evaluation metric.
Experiments indicate that, no matter what generator is employed, a sampling-insensitive discriminator, e.g. the commonly used PointNet-Max, will produce point cloud shapes with non-uniform density and clustering artifacts, while a sampling-oversensitive discriminator (\eg PointNet++, DGCNN) will lead to disastrous failures when adapting them to training point cloud GANs.
Interestingly, a simple PointNet-Mix baseline coupled with the most naive fully-connected generator achieves the best performance, indicating that the current bottleneck of point cloud GAN is on the discriminator side. For future discriminators, we suggest they should be more discriminative in shape and aware but not oversensitive to sampling patterns. For the sanity check of any novel discriminators, our proposed "no generator" experiment can be used.

\begin{ack}
This research is funded in part by Kuaishou Technology and supported by NSF grant IIS-1764078, NSF grant CHS-1528025, a grant from the SAIL-Toyota Center for AI Research, a Vannevar Bush Faculty Fellowship, and a gift from Amazon Web Services.
\end{ack}


\appendix
\section{Training Details}
For WGAN training, we set the gradient penalty coefficient $\lambda_{gp} = 1$. In each iteration, the discriminator gets updated ten times while the generator gets updated one time ($n_{critic} = 10$). The latent vector $z\in \mathbb{R}^{512}$ is sampled from a standard normal distribution $\mathcal{N}(0,\,I)$. We used the Adam optimizer for updating all the generator and discriminator networks with a learning rate ${10}^{-4}$ and other coefficients of $\beta_1 = 0.5$ and $\beta_2 = 0.999$. We train all the GANs for 6000 epochs on chair dataset and 1500 epochs on multi-category dataset. We observe all GANs converge at the end of training.

Our networks are implemented using PyTorch. We use one NVIDIA Titan-Xp to train a GAN model. We will release our code to facilitate research in this field.

\section{Network Architectures}
\subsection{Discriminators}
\paragraph{PointNet Discriminator}
For all our experiments, we use a fixed PointNet architecture as below. Note that mix pooling will double the feature dimension while max pooling and average pooling keep it unchanged. For each FC layer in the MLPs except the last one before the final output, we use LeakyRELU as the activation function.  To constrain the range of the discriminator output, we use a Sigmoid activation at the end, which we find helpful for stabilizing the training in our experiments. 

\begin{equation*}
\begin{aligned}
&\text{MLP}([3, 64, 128, 1024]) &\rightarrow&  ~~\text{Max/Avg/Mix-Pooling}() &\rightarrow& ~~C \in \mathbb{R}^{N\times 1024/1024/2048} \\
\rightarrow~~&\text{MLP}([1024/1024/2048, 512, 1]) &\rightarrow& ~~\text{Sigmoid}()
\end{aligned}
\end{equation*}

\paragraph{Attention-Max/Mix}
The network structures are shown below. 
\begin{equation*}
\begin{aligned}
&\text{MLP}([3, 32, 64]) & \rightarrow& F \in \mathbb{R}^{N\times 64} &\rightarrow & \text{MLP}([64, 32]) &   \rightarrow  G \in \mathbb{R}^{N\times 32}&\\
& & \rightarrow& F \in \mathbb{R}^{N\times 64} &\rightarrow & \text{MLP}([64, 32]) &   \rightarrow H\in \mathbb{R}^{N\times 32}&\\
&& \rightarrow& F \in \mathbb{R}^{N\times 64} &\rightarrow & \text{MLP}([64, 64]) &   \rightarrow K\in \mathbb{R}^{N\times 64}&\\
&F + \omega \text{SoftMax}(GH^{T})K & \rightarrow& \text{MLP}([64, 256, 1024]) & \rightarrow &  \text{Max/Mix-Pooling}() &&\\
& &\rightarrow&  \text{MLP}([1024/2048, 512, 1]) &\rightarrow& \text{Sigmoid}()&
\end{aligned}
\end{equation*}
$\omega \in \mathbb{R}$ is a learnable weight to balance between the original feature $F$ and the feature $GH^{T}K $ from self attention unit.

\paragraph{Original attention implementation in PU-GAN\cite{li2019pu}}
PU-GAN\cite{li2019pu} performs an additional early fusion after the first MLP, basically it performs max pooling to obtain $T  = \text{Max}(F) \in \mathbb{R}^{1\times 64}$ and  per-point concatenates $T$ to $F$ forming $F' = [F, \text{tile}(T)]$. Other than replacing $F$ by $F'$, it is almost same to our implementation. We modify the dimensionality of its final feature after the second max pooling to be 1024 for a fair comparison to other discriminators. Since this structure leverages two pooling layers, it is unfair to compare it with other discriminators and hard to see the effect of its self-attention unit, which is why we introduce our PointNet-Attention.

\paragraph{PointNet++}
To speed up the training, we use a slightly simplified version of PointNet++\cite{qi2017pointnet++} with the architecture shown below. Same to the original implementation, we use LeakyReLU and batch normalization for each FC layer in the set abstraction layers (SA), and only LeakyReLU for the FC layers in the final MLP.

\begin{equation*}
\begin{aligned}
&\text{SA}(512, 0.1, [3, 64, 64, 128]) &\rightarrow&  ~~\text{Max-Pooling}() \\
\rightarrow~~&\text{SA}(128, 0.2, [128+3, 128, 256, 256]) &\rightarrow&  ~~\text{Max-Pooling}()\\
\rightarrow~~& \text{GlobalSA}([256+3, 256, 512, 1024]) &\rightarrow& ~~\text{Max-Pooling}()\\
\rightarrow~~& \text{MLP}([1024, 512, 1]) &\rightarrow& ~~\text{Sigmoid}()
\end{aligned}
\end{equation*}

\paragraph{DGCNN}
To speed up the training, we use a slightly simplified version of DGCNN\cite{qi2017pointnet++} with the architecture shown below. Same to the original implementation, we use LeakyReLU and batch normalization for each convolutional layer in the EdgeConv layers.

\begin{equation*}
\begin{aligned}
&\text{EdgeConv}([6, 64]) &\rightarrow&~~  \text{Max-Pooling}()\\
\rightarrow~~&\text{EdgeConv}([64\times2, 64])   &\rightarrow&~~ \text{Max-Pooling}()\\
\rightarrow~~&\text{EdgeConv}([64\times2, 128]) &\rightarrow&~~  \text{Max-Pooling}()\\
\rightarrow~~&\text{EdgeConv}([128\times2, 256])  &\rightarrow&~~  \text{Max-Pooling}()\\ 
\rightarrow~~&\text{EdgeConv}([512, 1024])  &\rightarrow&~~  \text{Max-Pooling}()\\
\rightarrow~~& \text{MLP}([1024\times2, 512, 256, 1]) &\rightarrow&~~ \text{Sigmoid}()
\end{aligned}
\end{equation*}

\subsection{Generators}
We used the officially released code of TreeGAN\cite{shu20193d} and GraphCNN-GAN\cite{valsesia2018learning}. We implement the FC generator and the deformation generator with the architectures shown below. For each FC layer except the last one in the second MLP, we use LeakyReLU as the activation function. 

Note that our ground truth data are normalized point clouds with a zero center and a unit length diagonal. We hence use a Sigmoid activation at the last layer. And we translate the Sigmoid output by $(-0.5, -0.5, -0.5)$ to produce the final point cloud. 

\paragraph{FC Generator}
The structure of FC generator is shown below:
\begin{equation*}
\begin{aligned}
&\text{MLP}([512, 512, 512, 512, 2048, 2048\times3]) \rightarrow \text{Sigmoid}() - (0.5, 0.5, 0.5)
\end{aligned}
\end{equation*}

\paragraph{Deformation Generator}
In our experiments, we use a unit sphere as our template surface, because all of our real point clouds are sampled from closed surfaces. In addition to a latent code, we input 2048 uniformly sampled points on the unit sphere to our deformation generator. We randomly generate the points for each point cloud simply by normalizing random variables drawn from 3-dimension normal distribution into unit vectors\cite{muller1959note}. We add a batch normalization layer to the last FC layer in the first MLP, which we find important for its generation quality.
\begin{equation*}
\begin{aligned}
&\text{MLP}([3+512, 512, 512, 512, 512]) \rightarrow \text{MLP}([512, 64, 3])\rightarrow \text{Sigmoid}() - (0.5, 0.5, 0.5)
\end{aligned}
\end{equation*}

\section{More Results of PointNet-based Discriminator Variants}\label{sec:more_results}
In Table.\ref{table:more_results}, we provide more results of other variants of PointNet-based discriminators.
\paragraph{PointNet-Max-2048} Note that PointNet-Max discriminator generates a 1024-D feature after max-pooling while PointNet-Mix discriminator doubles the size to 2048-D due to the concatenation of max and average pooling features. This difference affects the weights of the following MLP layers. For a completely fair comparison, we also experiment with a PointNet-Max-2048 discriminator which has the following network structure:
$$\text{MLP}([3, 64, 128, 2048]) \rightarrow  \text{Max-Pooling}() \rightarrow
 \text{MLP}([2048, 512, 1]) \rightarrow \text{Sigmoid}().$$
To highlight the difference, in Table \ref{table:more_results}, we change the name of our previous PointNet-Max discriminator in the Table \ref{table:results} from Max to Max-1024. When paired with FC generator, comparing to PointNet-Max-1024, PointNet-Max-2048 shares a very similar performance. It gets slightly worse on the chair dataset except for on COV-EMD metric while outperforming PointNet-Max-1024 slightly on the multi-category dataset. Using both 2048-D feature, PointNet-Max-2048 is stil far behind PointNet-Mix, which again demonstrates the advantage of Mix-Pooling over Max-Pooling.
\paragraph{PointNet-Attention-Mix and original PU-GAN Discriminator\cite{li2019pu}}
In the Table \ref{table:results}, we show that PointNet-Attention can improve the performance on sampling-related metrics comparing to PointNet-Max, which means self-attention module does capture the point density distribution though its overall generation quality is worse than PointNet-Mix. Note that PointNet-Attention also leverages a max pooling before outputting scores, it would be interesting to know whether replacing the max pooling by mix pooling can further improve its performance. In Table.\ref{table:more_results}, we denote the two variants as Attention-Max and Attention-Mix, correspondingly. We find that PointNet-Attention-Mix actually performs worse than PointNet-Attention-Max. We argue that the mix-pooling's density awareness comes from average pooling, which computes the center of mass in the feature space and hence is aware to certain baised sampling, but self-attention module degrades the density awareness in the average feature. In PointNet-Attention-Mix, the mix pooling sees a weighted sum of the per-point feature $F$ and a blended feature $GH^{T}K$ from self-attention unit. Different with averaging, the learned correlation coefficients between the points $GH^{T}$ no longer maintains the information of point density distribution, and that's why leveraging a mix-pooling in this case doesn't help enforcing a more uniform point density.

Our implementation of PointNet-Attention is different with the original implementation in PU-GAN\cite{li2019pu}, which leverages two max pooling layers with one at an early stage and one at a later stage. We experiment with the original implementation in PU-GAN\cite{li2019pu} and find that it outperforms our Attention-Max/Mix however is still fall behind PointNet-Mix by a large margin.

\begin{table*}[h]
  \centering
  \scriptsize
  \begin{tabular}{c|c|c|c|c|c|c|c|c|c}
    \toprule
    {Dataset} & {Generator} & {Pooling} & {FPD-Mix $\downarrow$} & { FPD-Max$\downarrow$} & {FGD$\downarrow$}   & { MMD-E$\downarrow$} &  { MMD-C$\downarrow$} & {COV-E$\uparrow$} & {COV-C$\uparrow$}\\\hline
    \multirow{10}{*}{{Chair}}
    & {FC}             & Max-1024       & 1.571 & 0.211  & 7.030 & 0.1017 & 0.00164 & 23.56 & 72.75\\
    & {FC}             & \textbf{Max-2048}       & 1.638 & 0.224 & 7.926 & 0.1697 & 0.00144 & 42.26 & 70.98 \\
    & {FC}             & Mix       & \textbf{0.184} & \textbf{0.209}  & \textbf{2.124} & \textbf{0.0674} & 0.00196 & 73.64 & 74.96\\
    & {FC}             & Attention-Max & 0.635 & 0.672  & 4.971 & 0.1156 & 0.00160 & 68.92 & 70.54 \\
    & {FC}             & \textbf{Attention-Mix} & 0.759 & 0.814  & 5.532 & 0.1167 & 0.00167 & 69.36 & 68.04 \\
    & {FC}             & \textbf{Attention\cite{li2019pu}} & 0.582 & 0.602  & 4.945 & 0.1178 & 0.00164 & 70.98 & 71.72 \\
    & {Deform}    & Max-1024       & 0.913 & 0.268  & 5.602 & 0.0908 & 0.00201 & 68.5 & 72.61 \\ 
    & {Deform}    & Mix       & 0.534 & 0.373  & 2.836 & 0.0695 & 0.00200 & \textbf{76.29} & \textbf{75.11}\\ 
    & {Deform}    & Attention-Max & 0.696 & 0.755  & 2.987 & 0.1141 & 0.00160 & 68.77 & 69.36 \\
    & {Deform}    & \textbf{Attention-Mix} & 0.792 & 0.817  & 3.010 & 0.1141 & \textbf{0.00157} & 65.97 & 68.33 \\
    \midrule
    \multirow{9}{*}{{Multi}}
    & {FC}          & Max-1024       & 1.553 & 0.354 & 6.981 & 0.0842 & 0.00153 & 35.16 & 64.16\\ 
    & {FC}          & \textbf{Max-2048}  & 1.415 & 0.306 & 6.226 & 0.1253 & 0.00128 & 53.80 & \textbf{74.10}\\
    & {FC}          & Mix       & \textbf{0.255} & \textbf{0.285} & 2.550 & \textbf{0.0656} & 0.00184 & \textbf{73.5} & 72.16 \\
    & {FC}          & Attention-Max & 0.414 & 0.453 & 4.234 & 0.1188 & {0.00134} & 72.19 & 69.63 \\
    & {FC}             & \textbf{Attention-Mix} & 0.531 & 0.581  & 4.555 & 0.1135 & 0.00139 & 69.95 & 68.93 \\
    -Cat& {FC}             & \textbf{Attention\cite{li2019pu}} & 0.442 & 0.486  & 4.202 & 0.1191 & 0.001527 & 73.4 & 74.08 \\
    & {Deform} & Max-1024       & 1.072 & 0.633 & 3.845 & 0.0799 & 0.00179 & 62.5 & 64.5\\ 
    & {Deform} & Mix       & 0.614 & 0.349 & {2.451} & 0.0670 & 0.00191 & 70.83 & 68.83\\ 
    & {Deform} & Attention-Max & 0.616 & 0.720 & 2.531 & 0.1113 & 0.00141 & 72.04 & 69.60 \\
    & {Deform}    & \textbf{Attention-Mix} & 0.447 & 0.492  & \textbf {2.120} & 0.1085 & \textbf{0.00132} & 73.12 & 70.56 \\
    \hline
  \end{tabular}
  \vspace{-3mm}
  \caption{\textbf{Evaluating more variants of PointNet-based discriminators.}
  Here the new discriminators, Max-2048, Attention-Mix, and Attention\cite{li2019pu}, are in bold.}
  \label{table:more_results}
  \vspace{-3mm}
\end{table*}

\section{FPD Implementation}
We choose not to use the checkpoint of FPD metric in the released code\cite{treegan2019github} of TreeGAN\cite{shu20193d} but train our own FPD-Max/Mix. We found that, the released code \cite{treegan2019github} implements a PointNet-Max network containing a spatial transformer network for feature extraction. We empirically found that the spatial transformer network, which learns to rotate the real point clouds in ModelNet40\cite{wu20153d}, can lead to a very large variance in the FPD scores of generated point clouds. So, we remove the spatial transformer network from the PointNet feature extractor and stick to a vanilla PointNet in our FPD-Mix/Max implementation.

Here we compare the original FPD and our FPD-Max, both of which use PointNet-Max features of the generated point clouds from a deformation generator. This deformation generator is trained using a PointNet-Mix discriminator. We always evaluate FPD using a set of 10K samples. To see the variance between two closed checkpoints, we use one checkpoint at epoch 5990 and the other at epoch 6000 in this experiment. To obtain the variance of different sets of samples, we generate 5 sets of 10K samples for the checkpoint at epoch 6000 using different random seeds.

The comparison is shown in Table \ref{table:fpd_comparison}. Our FPD-Max has a lower relative standard deviation, which is defined as the ratio of the standard deviation to the absolute value of the mean. Also, given that our GAN is almost converged at epoch 6000, our FPD-Max only relatively changes $3.33\%$ from epoch 5990 to epoch 6000 while the original FPD changes $49.91\%$. The larger variance and the significant change indicate that the original FPD is not a stable metric for evaluating generated point clouds.

\begin{table*}[t!]
\centering
\scriptsize
\begin{tabular}{c|c|c|c|c|c|c}
\toprule
             & Mean@6000 & Std@6000 & Relative Std@6000 & Mean@5990 & Mean Diff & Relative Mean Diff \\\hline
Original FPD\cite{treegan2019github} & 3.814           & 0.2642         & 6.93\%                  & 1.910           & 1.9039    & 49.91\%            \\
FPD-Max      & 0.318           & 0.00655        & 2.06\%                  & 0.3082          & 0.0106    & 3.33\%             \\\hline
\end{tabular}
\caption{\textbf{Comparison between our FPD-Max and original FPD}: the first three columns show the mean, the standard deviation, and the relative standard deviation of the FPDs of the 5 different sets of generated samples using the checkpoint at epoch 6000. The fourth column shows the mean FPD of the generated samples using the checkpoint at epoch 5990. The firth and the sixth columns show the absolute and relative differences between the mean FPDs for checkpoint 6000 and 5990. }
\label{table:fpd_comparison}
\end{table*}

\vspace{-2mm}
\begin{figure*}[t!]
\centering
  \includegraphics[width=\linewidth]{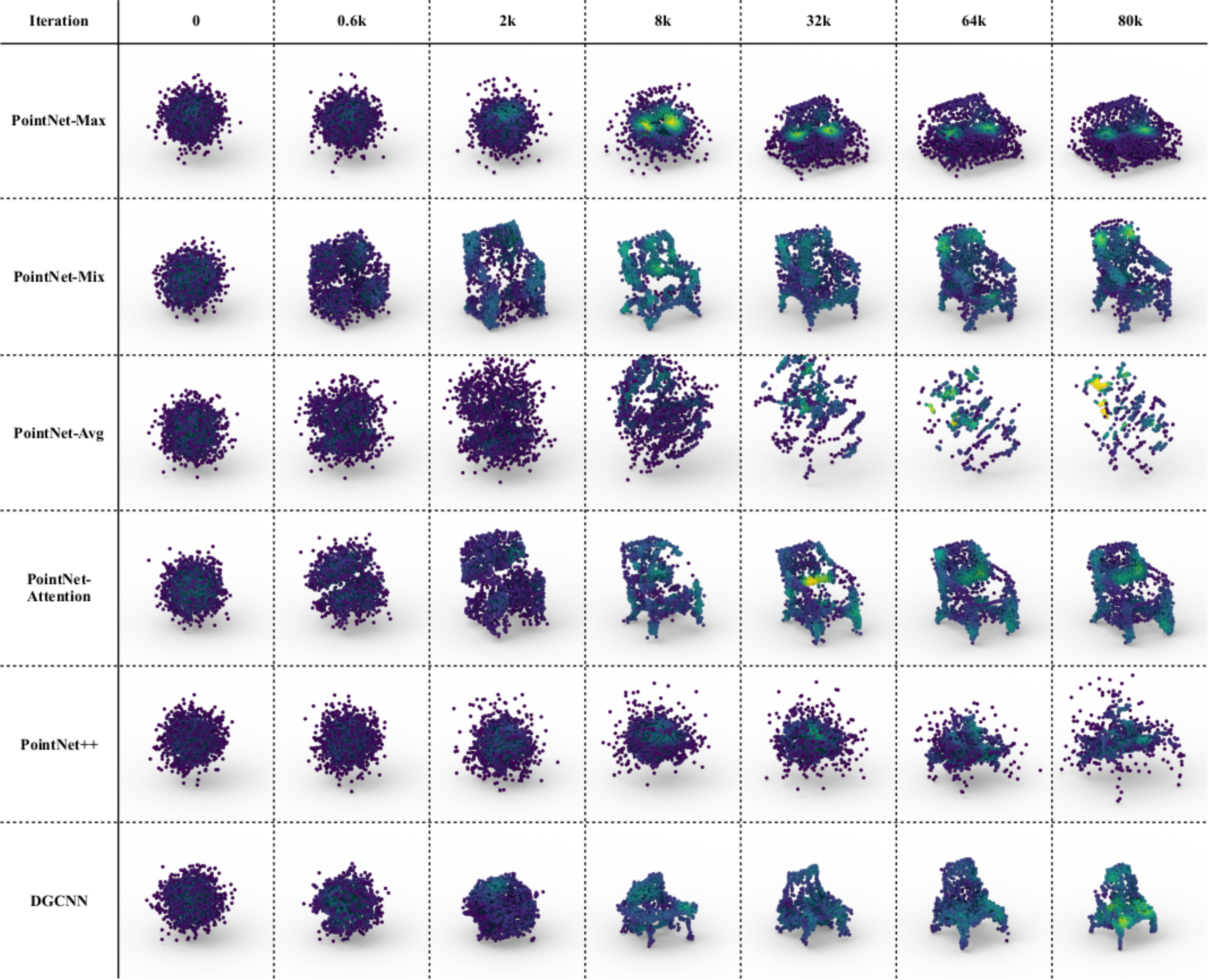}
  \caption{\textbf{Visualization of the evolution of one learnable point cloud for each discriminator during no generation trainings.} We color-code the local point density that ranges from sparse (dark blue) to dense (light yellow).} 
  \label{fig:no_generator_vis}
   \vspace{-3mm}
\end{figure*}

\section{Visualization of Learnable Point Clouds during No Generator Training}
In Figure \ref{fig:no_generator_vis}, we show the evolution of one learnable point cloud during its training process for each discriminator. The result indicates that only PointNet-Max/Mix/Attention are qualified as good teachers for point cloud generation. Compared to PointNet-Max, PointNet-Mix/Attention give gradients to every point, resulting in a faster convergence. Also, PointNet-Mix/Attention generates a more uniform shape comparing to PointNet-Max.

For point cloud density visualization used in Figure \ref{fig:no_generator_vis} and Figure \ref{fig:generation}, point densities are estimated using a linear kernel density estimation method with a bandwidth $0.1$. Basically, the local density of a point is proportional to the number of points within its ball neighbourhood with a radius of $0.1$. 

\clearpage
{\small
\bibliographystyle{ieee_fullname}
\bibliography{egbib}
}

\end{document}